\documentclass[10pt,twocolumn,letterpaper]{article}

\usepackage{cvpr}

\usepackage{booktabs}
\usepackage[utf8]{inputenc}
\usepackage[T1]{fontenc}
\usepackage{amsmath}
\usepackage{amssymb}
\usepackage[ruled,vlined]{algorithm2e}
\usepackage{microtype}
\usepackage{multirow}
\usepackage[utf8]{inputenc}
\usepackage[T1]{fontenc}
\usepackage{booktabs}
\usepackage{multirow}
\usepackage{tabularx}

\definecolor{cvprblue}{rgb}{0.21,0.49,0.74}
\usepackage[pagebackref,breaklinks,colorlinks,allcolors=cvprblue]{hyperref}

\def\paperID{*****}
\def\confName{CVPR}
\def\confYear{2025}

\title{Technical Report for Argoverse2 Scenario Mining Challenges on Iterative Error Correction and Spatially-Aware Prompting}

\author{Yifei Chen\\
Xi'an University of Technology\\
{\tt\small yfchen@stu.xaut.edu.cn}
\and
Ross Greer\\
University of California, Merced\\
{\tt\small rossgreer@ucmerced.edu}}

\begin{document}
\maketitle

\begin{abstract}
Scenario mining from extensive autonomous driving datasets, such as Argoverse~2, is crucial for the development and validation of self‑driving systems. The RefAV framework represents a promising approach by employing Large Language Models (LLMs) to translate natural‑language queries into executable code for identifying relevant scenarios. However, this method faces challenges, including runtime errors stemming from LLM‑generated code and inaccuracies in interpreting parameters for functions that describe complex multi‑object spatial relationships. This technical report introduces two key enhancements to address these limitations: (1)~a fault‑tolerant iterative code‑generation mechanism that refines code by re‑prompting the LLM with error feedback, and (2)~specialized prompt engineering that improves the LLM's comprehension and correct application of spatial‑relationship functions. Experiments on the Argoverse~2 validation set with diverse LLMs—Qwen2.5‑VL-7B, Gemini 2.5 Flash, and Gemini 2.5 Pro—show consistent gains across multiple metrics; most notably, the proposed system achieves a HOTA‑Temporal score of \textbf{52.37} on the official test set using Gemini 2.5 Pro. These results underline the efficacy of the proposed techniques for reliable, high‑precision scenario mining.
\end{abstract}

\section{Introduction}
The deployment of Autonomous Vehicles (AVs) necessitates rigorous testing and validation, for which the identification of interesting, rare, or safety-critical scenarios from vast operational data is paramount. This process is vital not only for evaluating ego-behavior and safety testing but also for enabling active learning at scale \cite{greer2025language}. Traditional methods relying on manual inspection or predefined heuristics are often prohibitively time-consuming and prone to errors when faced with terabytes of multi-modal data collected by AV fleets \cite{8906293}. Previous methods that used database queries for scenario mining lacked flexibility compared to methods that used LLM \cite{Gelder_2020, haussmann2020scalable, keskar2025evaluating}. The huge volume and complexity of this data pose a challenge, making efficient and accurate scene mining a major ongoing challenge.

To address this, the Argoverse 2 Scenario Mining Challenge \cite{davidson2025refav,wilson2023argoverse2generationdatasets} provides a standardized benchmark, featuring 10,000 planning-centric natural language queries designed to retrieve specific scenarios from sensor data. The RefAV framework has emerged as a notable baseline for this challenge\cite{davidson2025refav}, leveraging the power of Large Language Models (LLMs). RefAV converts a natural language description of a scene into composable function calls; that is, it builds an executable script by chaining together calls to a pre-defined library of atomic functions, where each function represents a simple action such as finding an object with a directional relationship or a nearby object, and includes basic AND, OR, and NOT operations. This approach offers flexibility and expressiveness beyond structured query languages. For example, a query like, 'a truck cuts off a car, causing the car to brake hard,' would be difficult to formulate in a rigid DSL that lacks operators for causality and sequencing. This method, however, allows the LLM to generate a script that identifies a truck's lane change event and then searches for a subsequent hard-braking event from a nearby car within a short time window, effectively capturing the causal link implied by the natural language.

Despite the promise of LLM-based scenario mining, practical implementations like RefAV encounter specific limitations. Firstly, code generated directly by LLMs can frequently contain syntactic or logical errors, leading to runtime failures. These failures disrupt the mining pipeline and result in incomplete scenario discovery. Secondly, LLMs may struggle with the nuanced semantics of functions describing relative spatial relationships between multiple objects. For instance, functions such as \textit{has objects in relative direction()} or \textit{facing toward()} require precise parameter assignment to reflect the intended meaning (e.g., distinguishing "a car in front of a pedestrian" from "a pedestrian in front of a car"). When the track candidates parameter is a pedestrian, the related candidates parameter is a car, and the direction is forward, the function correctly represents “a car in front of a pedestrian,” but LLM often reverses the track candidates and related candidates. Misinterpretation of these parameters leads to semantic inaccuracies in the retrieved scenarios, even if the code executes without error. This issue is a manifestation of a known failure mode in LLMs, often termed `factual hallucination' or a breakdown in understanding relational knowledge \cite{li-etal-2024-dawn,rawte2023surveyhallucinationlargefoundation}. These are not merely superficial issues but represent fundamental hurdles in reliably converting complex human language into precise and correct machine-executable instructions.

\section{Method}
This section first provides an overview of the baseline RefAV scenario mining pipeline. Subsequently, it details the two primary contributions of this work: the Fault-Tolerant Iterative Code Generation (FT-ICG) mechanism and the Enhanced Prompting for Spatial Relational Functions (EP-SRF).
\begin{figure*}
    \centering
    \includegraphics[width=1\linewidth]{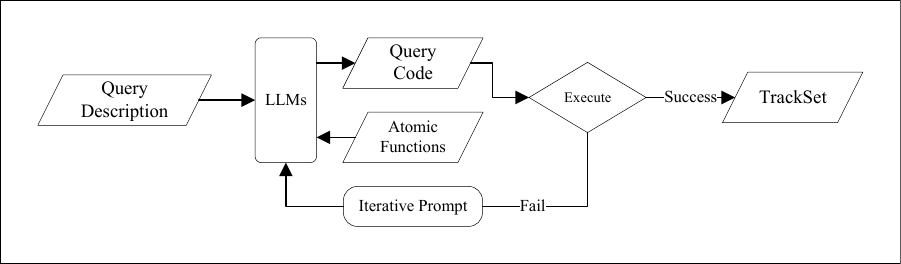}
    \caption{Flowchart of Fault-Tolerant Iterative Code Generation}
    \label{fig:enter-label}
\end{figure*}
\subsection{Fault-Tolerant Iterative Code Generation (FT‑ICG)}
A significant challenge in the practical application of LLMs for code generation is the propensity for the generated code to contain errors. These errors can range from simple syntax mistakes to more complex logical flaws or incorrect usage of the provided atomic functions, all of which lead to runtime exceptions. Such failures can terminate the scenario mining process prematurely, resulting in missed scenarios and reduced overall system reliability.
\begin{algorithm}[h]
  \caption{Fault-Tolerant Iterative Code Generation}
    \label{algorithm1}
  \KwIn{Natural-language query $\mathit{NLQuery}$; set of atomic functions $\mathcal{A}$; maximum iterations $K$}
  \KwOut{Executable Python code $\mathit{ValidCode}$}
  \BlankLine

  \SetKw{Try}{try}
  \SetKw{Catch}{catch}
  \SetKw{Break}{break}
  \SetKw{RuntimeError}{RuntimeError}

  $Prompt \leftarrow \textsc{Compose}\bigl(\mathit{NLQuery},\ \textsc{Describe}(\mathcal{A})\bigr)$\;

  \For{$i \gets 1$ \KwTo $K$}{
    \Try{
      \Indp\\
        $Code \leftarrow \textsc{LLMGenerate}(Prompt)$\;
        \textsc{PythonExec}$(Code)$\;
         Break;\\
      \Indm
    }%
    \Catch(\RuntimeError $\varepsilon$){
      \Indp
        $ErrorMsg \leftarrow \textsc{Message}(\varepsilon)$\;

        $IterationPrompt \leftarrow$ \texttt{%
        "This is the code generated last time: }\,$\{Code\}$%
        \texttt{, with the error message: }\,$\{ErrorMsg\}$%
        \texttt{. Please avoid code runtime errors."}\;

        $Prompt \leftarrow \textsc{Compose}\bigl(\mathit{NLQuery},\ IterationPrompt\bigr)$\;
      \Indm
    }
  }
\end{algorithm}
The pseudocode for the fault-tolerant iterative code generation mechanism is shown in Algorithm \ref{algorithm1}. This iterative approach treats the LLM not as a single-shot code generator but as an entity capable of learning from explicit feedback on its errors. By providing the context of the previous failure, the LLM is guided towards a correct solution. This significantly increases the success rate of code generation, thereby enhancing the robustness and coverage of the scenario mining pipeline, allowing it to handle a broader spectrum of queries and code complexities without manual intervention. This process mirrors a human programmer's debugging cycle, iteratively refining code based on observed errors.

\subsection{Enhanced Prompting for Spatial Relational Functions (EP-SRF)}
Beyond syntactic correctness, the semantic accuracy of the generated code is paramount. LLMs often fail to correctly interpret and parameterize functions that describe the relative spatial relationships between multiple objects in a specific domain. For example, a query like "a cyclist to the left of a bus" requires the LLM to correctly assign the `cyclist' and `bus' tracks to the appropriate parameters of a function like has \textit{objects in relative direction()}. An incorrect assignment could lead the system to search for "a bus to the left of a cyclist," fundamentally misinterpreting the query.
To mitigate such semantic errors, Enhanced Prompting for Spatial Relational Functions is introduced. This involves augmenting the initial prompt provided to the LLM with specific instructions that clarify the argument semantics for these critical functions. Before the LLM attempts to generate code involving functions that define relative positions or orientations, it receives the following guiding information:

\textit{If you use} has objects in relative direction(), being crossed by(), heading in relative direction to() \textit{functions, direction parameter specifies the orientation of related candidates relative to track candidates. The} facing toward() \textit{and} heading toward() \textit{functions indicate that the track candidates parameter is oriented toward the related candidates parameter.}

This explicit instruction serves as a form of contextual disambiguation. It clearly defines the roles of \textit{track candidates} (often the primary subject of the relation) and \textit{related candidates} (the reference object) within the context of each specified function. For directional functions like \textit{has objects in relative direction}, it clarifies which entity's perspective defines the direction. For orientational functions like \textit{facing toward}, it specifies which entity is performing the action of facing. By providing this upfront clarification, the LLM is better equipped to map the natural language description of spatial relationships to the correct functional representation and parameter assignment. This leads to a higher fidelity in translating complex spatial queries, ultimately improving the semantic accuracy and relevance of the mined scenarios. This addresses the challenge that code might run correctly but perform the wrong semantic operation if the LLM misunderstands these subtle but critical distinctions.
\section{Experiments}
In this section, we provide some experimental details for reproducibility of the final results, which were evaluated on ArgoVerse2.
\subsection{Implementation Details}
The experiments were conducted using the Argoverse 2 dataset. The dataset provides rich multi-modal information, including RGB camera frames, LiDAR point clouds, HD Maps, and 3D track annotations for 26 object categories. 

The primary metric is HOTA-Temporal. It is a spatial tracking metric that considers only the scenario-relevant objects during the precise timeframe when the scenario is occurring. 
HOTA\cite{Luiten_2020} was introduced to provide a unified evaluation of multi-object tracking by jointly accounting for detection, association, and localization—three facets that together reflect human intuition of tracking quality.
Secondary metrics include HOTA, Timestamp F1, and Log F1. Timestamp F1 treats the video as a sequence of frames, labeling each timestamp as “scenario” or “non-scenario.” Precision and recall are computed from the comparison of predicted and ground-truth frame labels. Log F1 simplifies the task to a single binary decision per log. After aggregating true positives, false positives, and false negatives across all logs, a conventional F1-score is produced. 

In our setup, the Qwen2.5-VL-7B model \cite{bai2025qwen25vltechnicalreport} was deployed locally on a workstation outfitted with an NVIDIA RTX 4090 GPU, whereas the Gemini model \cite{geminiteam2025geminifamilyhighlycapable} was accessed remotely via API calls. For 3D object detection and tracking, we utilized the Le3DE2D track obtained directly from the official LT3D method \cite{peri2023longtailed3ddetection}. We set \textit{K} in Algorithm \ref{algorithm1} to 5. For the generated code, if the number of iterations of the fault tolerance mechanism exceeds the \textit{K} value, we manually edit the generated code, manually modify the reported errors, and fill in the correct track candidates, related candidates, and direction parameters.

\subsection{Ablation Studies}
Ablation studies were performed on the Argoverse 2 validation set to systematically assess the individual and combined effects of the Fault-Tolerant Iterative Code Generation (FT-ICG) and Enhanced Prompting for Spatial Relational Functions (EP-SRF).

\begin{table}
  \centering
  \resizebox{0.5\textwidth}{!}{%
    \begin{tabular}{llcccc}
      \toprule
      Model & Method & HOTA-T & HOTA & TS-F1 & Log-F1 \\
      \midrule
      \multirow{3}{*}{Qwen2.5-VL-7B}
        & Baseline RefAV      & 33.27 & 36.72 & 61.94 & 58.12 \\
        & + FT-ICG            & 34.71 & 39.32 & 62.77 & 58.09 \\
        & + FT-ICG + EP-SRF   & \textbf{37.55} & \textbf{42.48} & \textbf{65.03} & \textbf{60.90} \\
      \midrule
      \multirow{3}{*}{Gemini 2.5 Flash}
        & Baseline RefAV      & 42.73 & 44.27 & 69.84 & 60.13 \\
        & + FT-ICG            & 44.13 & 45.07 & 70.44 & 60.66 \\
        & + FT-ICG + EP-SRF   & \textbf{44.58} & \textbf{45.12} & \textbf{71.54} & \textbf{60.79} \\
      \midrule
      \multirow{3}{*}{Gemini 2.5 Pro}
        & Baseline RefAV      & 43.34 & 45.57 & 69.84 & 59.13 \\
        & + FT-ICG            & 45.53 & 46.07 & 71.34 & 59.66 \\
        & + FT-ICG + EP-SRF   & \textbf{46.71} & 45.93 & \textbf{72.30} & \textbf{61.36} \\
      \midrule
      \multirow{1}{*}{Gemini 2.5 Pro (Test)}
        & + FT-ICG + EP-SRF   & \textbf{52.37} & \textbf{51.53} & \textbf{77.48} & \textbf{65.82} \\
      \bottomrule
    \end{tabular}%
  }
  \caption{Validation results of different methods on Qwen2.5-VL-7B, Gemini 2.5 Flash and Gemini 2.5 Pro.}
  \label{tab:merged}
\end{table}

Across all three LLMs, the FT-ICG mechanism consistently improves performance, particularly in HOTA-Temporal. This underscores the practical benefit of addressing runtime code errors. The subsequent addition of EP-SRF generally provides further enhancements, especially in HOTA-Temporal, Timestamp F1, and Log F1, highlighting the importance of semantic accuracy in function parameterization. The consistency of these gains across diverse LLMs suggests that the proposed methods address fundamental challenges in LLM-based code generation and interpretation rather than model-specific idiosyncrasies.

The results on the test set are also strong, with a HOTA-Temporal score of 52.37. This performance is significantly higher than the validation set results for the same configuration (46.71 HOTA-Temporal), which is a positive indication of generalization and potentially reflects differences in data distribution or complexity between the validation and test splits. The scores across all metrics confirm the effectiveness of the combined approach.

FT‑ICG raises robustness by recovering from syntax and logic errors, improving recall. EP‑SRF sharpens semantic precision, boosting Timestamp and Log‑level metrics. Even the strongest LLM gains, indicating that domain‑specific scaffolding complements raw model capability.

\section{Conclusion}
The Fault-Tolerant Iterative Code Generation and Enhanced Prompting for Spatial Relational
Functions enhancements to robustness make LLM-driven scene mining on Argoverse 2 more fault-tolerant and semantically accurate. The methods deliver state‑of‑the‑art HOTA‑Temporal scores without manual intervention. Future work includes dynamic prompt adaptation, tighter integration of multimodal cues, and exploration of alternative latent representations of scenes for novelty mining \cite{greer2024towards}.

{\small
\bibliographystyle{ieeenat_fullname}
\bibliography{main}
}

\end{document}